\documentclass[letterpaper, 10 pt, conference]{ieeeconf}  

\IEEEoverridecommandlockouts                              

\usepackage{graphicx} 
\usepackage{amsmath} 
\usepackage{amssymb}  
\usepackage[nocompress]{cite}
\usepackage{bbm}
\usepackage{makecell}
\usepackage{caption}
\usepackage{subcaption}

\makeatletter
\let\NAT@parse\undefined
\makeatother
\usepackage[pagebackref,breaklinks,colorlinks]{hyperref}

\usepackage[capitalize]{cleveref}
\crefname{section}{Sec.}{Secs.}
\Crefname{section}{Section}{Sections}
\Crefname{table}{Table}{Tables}
\crefname{table}{Tab.}{Tabs.}

\newcommand{\blindSwitch}[2]{#1}
\newcommand{\LV}{\blindSwitch{LV}{A}}
\newcommand{\SEA}{\blindSwitch{SEA}{B}}
\newcommand{\SF}{\blindSwitch{SF}{C}}
\newcommand{\SLAC}{\blindSwitch{SLAC}{D}}

\title{\LARGE \bf
Generating Driving Scenes with Diffusion
}

\author{Ethan Pronovost\textsuperscript{1}, Kai Wang\textsuperscript{1}, Nick Roy\textsuperscript{1, 2}
\thanks{\textsuperscript{1}Zoox. Contact: {\tt\small epronovost@zoox.com}}%
\thanks{\textsuperscript{2}CSAIL, Massachusetts Institute of Technology (MIT).}%
}

\begin{document}

\maketitle
\thispagestyle{empty}
\pagestyle{empty}

\begin{abstract}

In this paper we describe a learned method of traffic scene generation designed to simulate the output of the perception system of a self-driving car. In our ``Scene Diffusion’’ system, inspired by latent diffusion, we use a novel combination of diffusion and object detection to directly create realistic and physically plausible arrangements of discrete bounding boxes for agents. We show that our scene generation model is able to adapt to different regions in the US, producing scenarios that capture the intricacies of each region. 

\end{abstract}

\section{INTRODUCTION}

\begin{figure*}[t]
     \centering
     \begin{subfigure}[b]{0.3\textwidth}
         \centering
         \includegraphics[width=\textwidth]{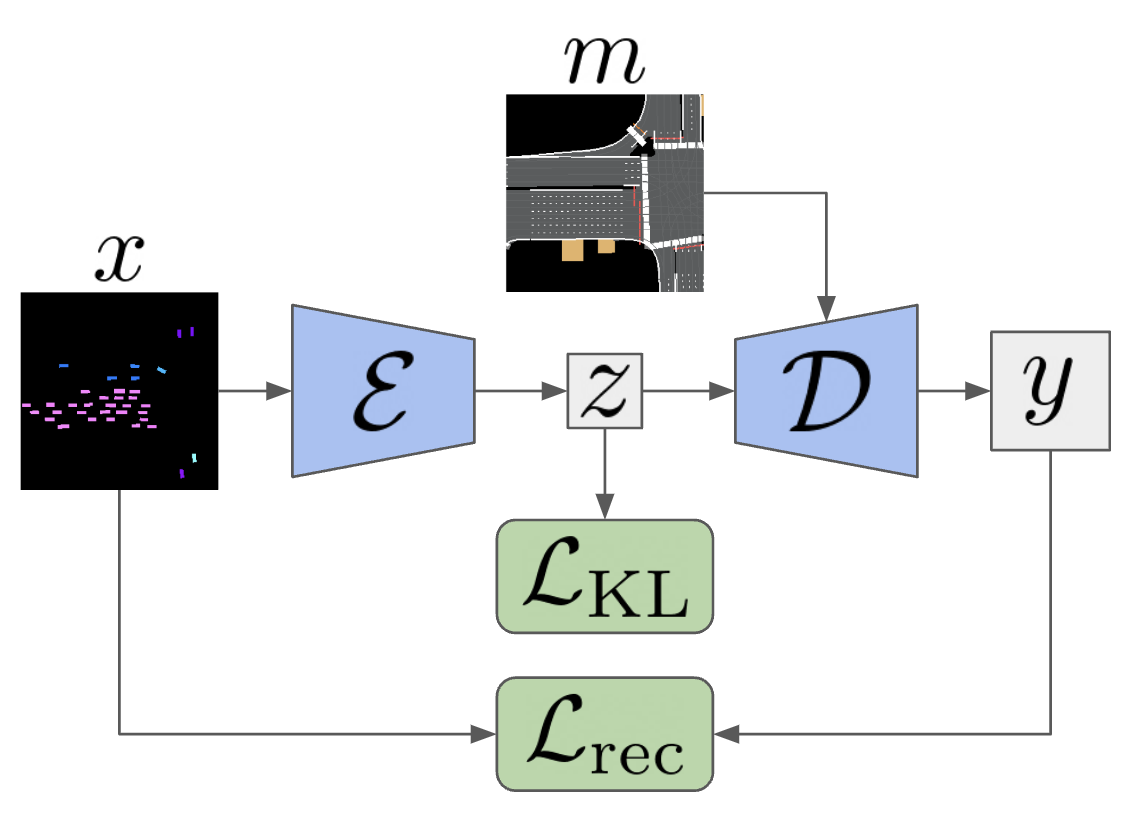}
         \caption{Autoencoder training.}
         \label{fig:autoencoder_architecture}
     \end{subfigure}
     \hfill
     \begin{subfigure}[b]{0.38\textwidth}
         \centering
         \includegraphics[width=\textwidth]{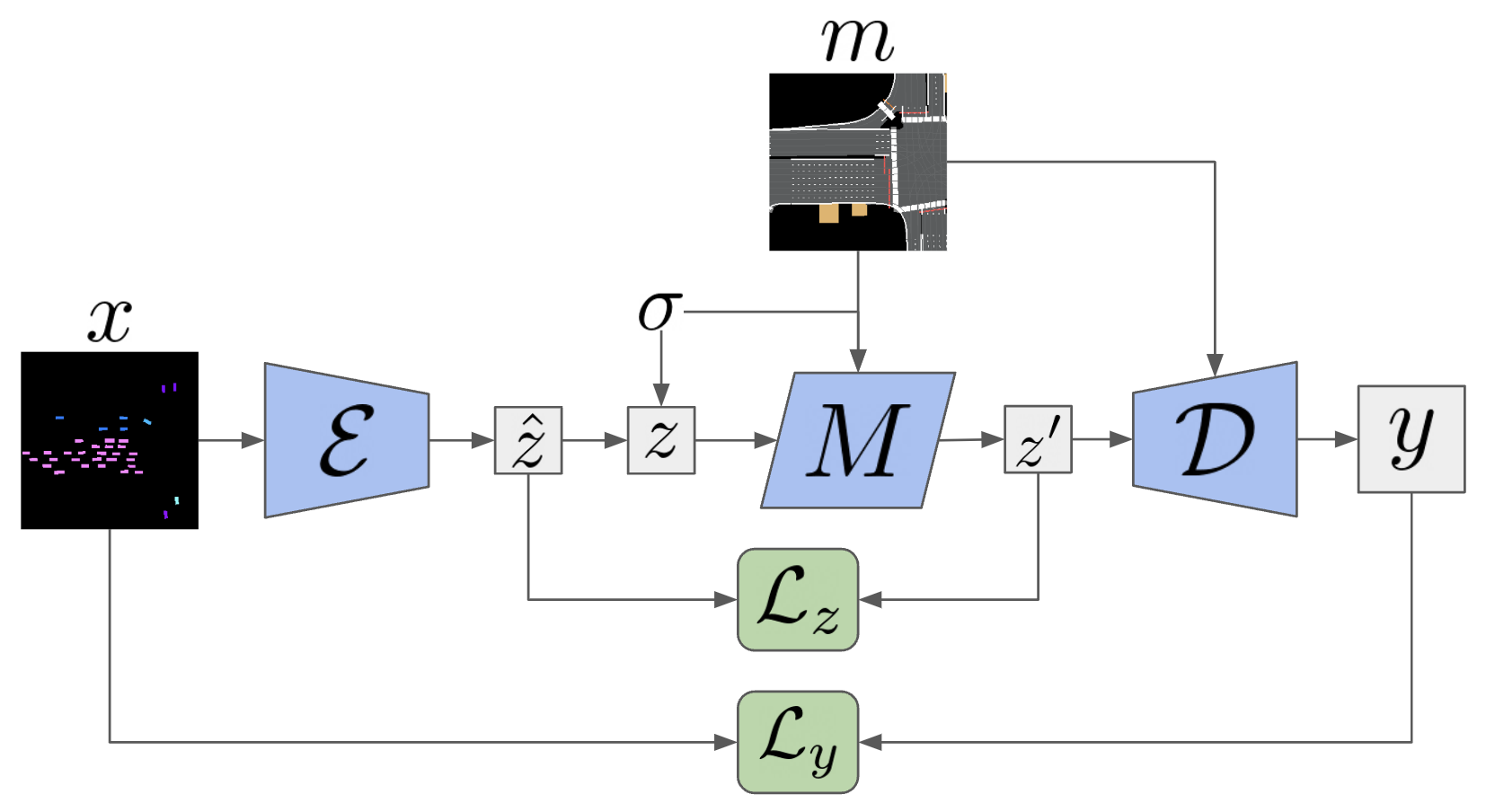}
         \caption{Diffusion training.}
         \label{fig:diffusion_architecture}
     \end{subfigure}
     \hfill
     \begin{subfigure}[b]{0.3\textwidth}
         \centering
         \includegraphics[width=\textwidth]{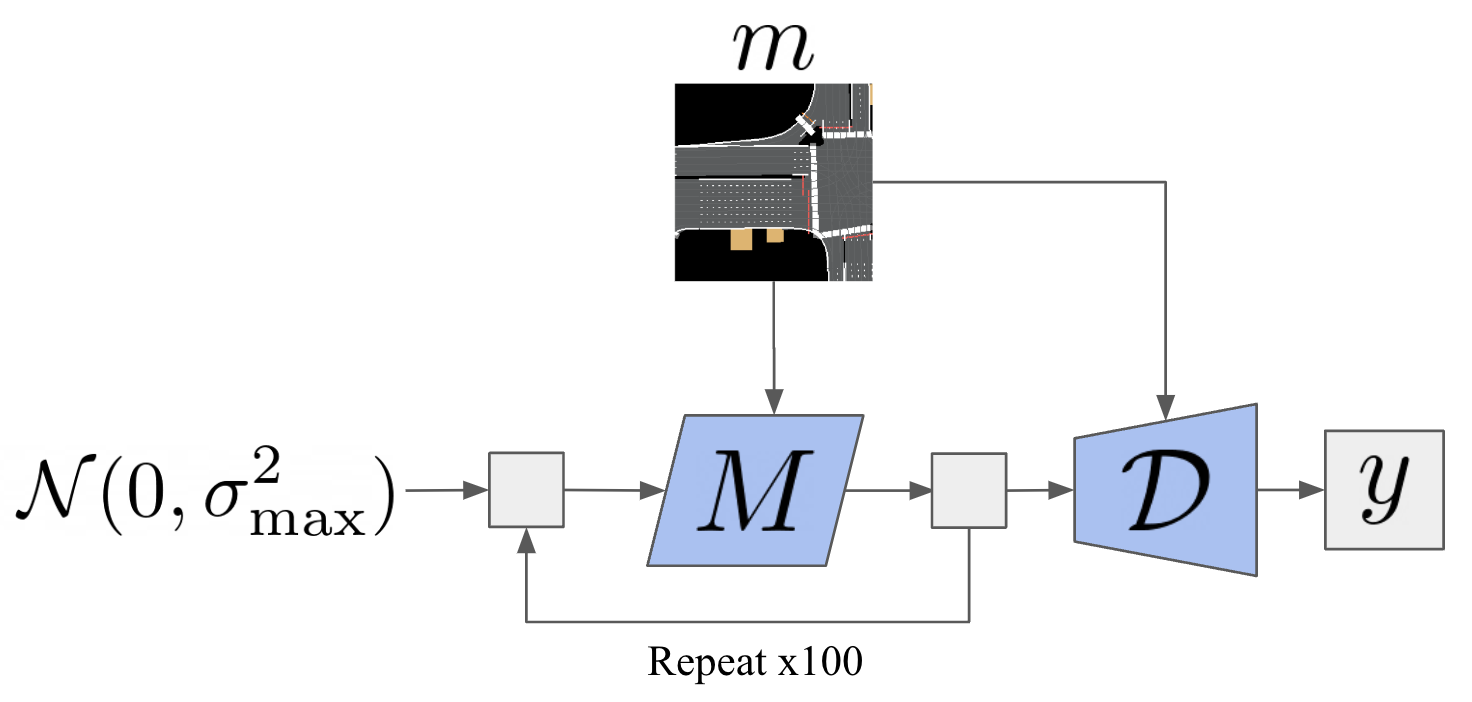}
         \caption{Inference.}
         \label{fig:inference_architecture}
     \end{subfigure}
    \caption{Architectures for training and inference. In (a) an autoencoder is trained to encode a birds' eye view image of vehicles in a scene ($x$) and output oriented bounding box detections ($y$) for the entities. In (b) the pre-trained autoencoder is used to train a diffusion model on the latent embeddings ($z$) of the autoencoder conditioned on a map image ($m$). In (c) the diffusion model and decoder are used to generate novel traffic scenes by first running diffusion inference in the latent space and then decoding to recover oriented bounding boxes.}
    \label{fig:architectures}
\end{figure*}

Simulation has long been a useful component for integration testing as part of the development of autonomous vehicles. The advent of high quality photorealism and realistic physics in recent simulators has enabled the development and evaluation of new models and algorithms for autonomy with much more reliability than has historically been possible \cite{Brooks1993}. 

However, one of the limitations of all simulations is the difficulty in creating a range of simulated scenarios that vary in ways that accurately match the distribution of scenarios in the real world. Given a set of simulation assets, those assets must still be arranged in a scenario that is physically plausible, e.g., for a traffic simulator, simulated vehicles must be oriented correctly with respect to the road surface and their motion must be a reasonable facsimile of human driving. Scenario construction can be performed by hand or with hand-crafted heuristics, but the number of scenarios that can be constructed manually is limited. Unusual (but plausible) maneuvers such as cars moving into oncoming traffic around a double-parked vehicle may not be easily captured with heuristics. 

For the purposes of developing prediction and planning algorithms for an autonomous vehicle, we are less interested in photorealistic scenario generation than we are in simulating an abstraction of the scenario that would be produced by a perception system such as dynamic, oriented bounding boxes that represent cars and pedestrians in the environment.
Several recent results in procedural scene generation have shown promise in learning different models of the distributions of real-world scenes. However, the computational and data complexity of learning explicit statistical models such as scene grammars tend to be restricted to small scale, static scenes \cite{izatt2020generative,fisher2012example}. More recent deep learning models have been shown to produce complex simulation scenes with larger numbers of agents \cite{scenegen_2021, trafficgen_2022, simnet_2021}. These works are based on established techniques for image generation  \cite{pixelrnn_2016, cgan_2016}.

The development of diffusion models \cite{ddpm_2020} has shown to be remarkably successful in image generation, surpassing prior methods \cite{diffusion-beats-gans_2021}.
This work is motivated by the hypothesis that diffusion is also a better approach for generating traffic scenes.
Latent diffusion \cite{latent-diffusion_2022} is especially well suited for generating traffic scenes due to the decoupling of latent and output spaces.
However, the image-based loss term that is conventionally used in generative models such as autoencoders needs to be adapted to this kind of abstract scene output. 

In this paper we describe a traffic scene generation architecture we refer to as ``Scene Diffusion’’.
Following \cite{latent-diffusion_2022}, there are two parts to our model architecture: an autoencoder which is trained first, and a diffusion model which is trained second on the latent embeddings from the autoencoder.
We use a novel combination of diffusion and object detection to directly output discrete bounding boxes for agents.

The contributions of our work are:
\begin{itemize}
    \item We propose a novel end-to-end differentiable architecture based on latent diffusion and object detection for generating driving scenes.
    \item We evaluate the generalization capabilities of our scene generation model across different geographical regions qualitatively and quantitatively.
\end{itemize}

\section{PROBLEM SETTING} \label{sec:problem_setting}

Our goal is to develop a generative model to produce driving scenes conditioned on map data.
In this work a driving scene consists of a map and a set of agents, where each agent is described by an oriented bounding box.

The map is represented as a multi-channel birds' eye view image $m \in \mathbb{R}^{C_m \times H \times W}$ similar to that described in \cite{scenegen_2021, rules-of-road_2019}.
This map image contains information about the road geometry, regions of interest (e.g. driveways, crosswalks, parking spots), and traffic light states.

The agents are also represented as a multi-channel birds' eye view image $x \in \mathbb{R}^{C_x \times H \times W}$, also similar to \cite{scenegen_2021, rules-of-road_2019}.
In addition to a binary channel representing whether a pixel is occupied we also use channels that fill each agent's bounding box with the sine and cosine of the agent's heading to disambiguate the orientation of each box.
This image $x$ is used to represent scenes during training and is not used for inference.
An example of $x$ and $m$ is shown in \cref{fig:architectures}.

At inference we wish to generate a set of bounding boxes conditioned on the map data $m$.
While these boxes are eventually converted to a center pose, length, and width, the model itself can use a different parameterization as described in \cref{sec:oriented_bbox}.

\section{SCENE AUTOENCODER} \label{sec:autoencoder}

The architecture of the scene autoencoder is adapted from that of \cite{latent-diffusion_2022}.
The main modification is in the decoder output representation and the associated reconstruction loss.
The architecture for training the autoencoder is shown in \cref{fig:autoencoder_architecture}.

\subsection{Conditional Variational Autoencoder} \label{sec:cvae}

The agent image $x$ is processed by an encoder $\mathcal E(x)$ to obtain a mean and standard deviation $z_\mu, z_\sigma \in \mathbb{R}^{C' \times H' \times W'}$, where $H' = H/2^f$ and $W' = W/2^f$ for some down-sampling count $f \in \mathbb{N}$.

From these parameters, a latent embedding is obtained using the reparametrization trick \cite{kingma_variational-bayes_2013}: $z = z_\mu + z_\sigma \epsilon$ where $\epsilon \sim \mathcal N \left( 0, \mathbf I \right)$.
This latent embedding is then given to a decoder $\mathcal{D}$ along with $m$.
The decoder output is $y = \mathcal{D}(z; m)$.

To train the VAE, a reconstruction loss $\mathcal L_\textrm{rec} (y, x)$ is applied between the decoder output and the original encoder inputs $x$.
A KL regularization loss is applied to the latent embedding distributions, 
\begin{equation}
\mathcal L_\textrm{KL} = D_{KL} \left[ \mathcal N(z_\mu, z_\sigma^2) \, || \, \mathcal N(0, \mathbf{I}) \right]
\end{equation}

The final loss for the VAE is a weighted combination of these two losses \cite{higgins_beta-vae_2022}:
\begin{equation}
\mathcal L_\textrm{VAE} = \mathcal L_\textrm{rec} + \beta_\textrm{KL} \mathcal L_\textrm{KL}
\end{equation}

One of the key adaptations to go from images to driving scenes is to change the output representation of the decoder and the reconstruction loss to train it.
The exact structure of $y$ and $\mathcal L_\textrm{rec}$ is described in the sections below.

\subsection{Oriented Bounding Boxes} \label{sec:oriented_bbox}

We aim to produce an anchor-free one-to-one object detector \cite{sun_what_2021} that will produce a single oriented bounding box per agent without needing heuristic post-processing to construct boxes as in \cite{simnet_2021, oriented-bbox-rotated_2021}.
However, the approach of \cite{sun_what_2021} is not immediately applicable to \emph{oriented} bounding box detection because differentiable IoU loss is not tractable for rotated bounding boxes.

Our simplified detection problem setting (we provide perfectly rendered boxes with channels clearly indicating the correct orientation of the box) allows us to use a more straight-forward approach compared to prior works performing oriented object detection in more challenging settings \cite{r3det_2019,  gaussian-wasserstein-detection_2021,  oriented-bbox-rotated_2021}.

\subsubsection{Representation}
In the decoder output $y$, each pixel represents one bounding box. Note that the pixel dimensions of $y$ can be different from those of $x$; in practice we use a lower resolution for $y$ to reduce computational and memory requirements.

Let $b$ be the feature vector for one pixel in $y$. The pixel location defines the 2D reference point $r(b) \in \mathbb{R}^2$ used to locate the box in space. 
There are seven channels of $b$ to define the probability, shape, and orientation of the box: $(l, \theta_c, \theta_s, d_\textrm{front}, d_\textrm{left}, d_\textrm{back}, d_\textrm{right})$

The first channel $l$ is a sigmoid logit that defines the probability for the box. 
The next two channels represent the cosine ($\theta_c$) and sine ($\theta_s$) of the box orientation.
By predicting sine and cosine values we avoid having an arbitrary discontinuity in the expected model output (e.g. between $-\pi$ and $\pi$).

\begin{figure}[t] 
\centering
\includegraphics[width=0.5\linewidth]{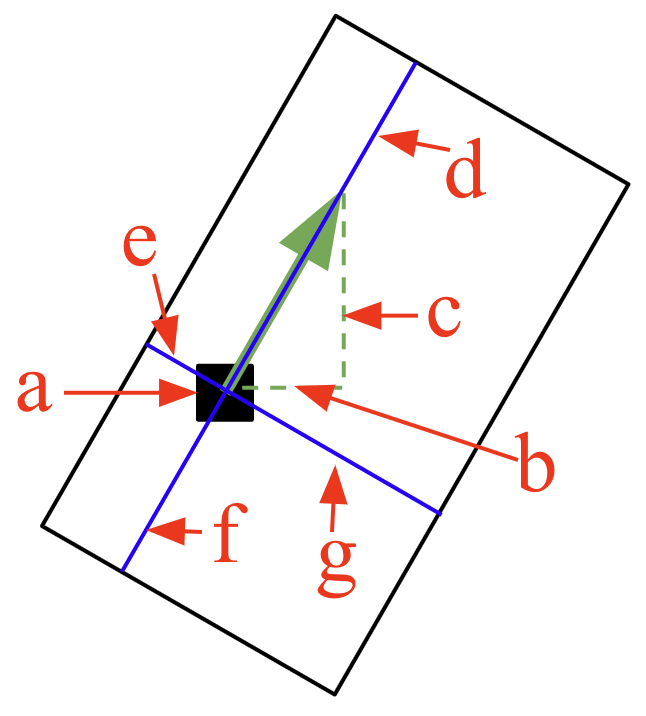}
\caption{Depiction of the bounding box parametrization. (a) shows the reference pixel that defines $r(b)$. (b) shows $\theta_c$ and (c) shows $\theta_s$, which combine to define the heading of the box. (d-g) show $\exp(d_\textrm{front})$, $\exp(d_\textrm{left})$, $\exp(d_\textrm{back})$, and $\exp(d_\textrm{right})$, respectively.}
\label{fig:box_params}
\end{figure}

Four more channels represent the log of the distances from the pixel center to the front ($d_\textrm{front}$), left ($d_\textrm{left}$), back ($d_\textrm{back}$), and right ($d_\textrm{right}$) sides of the box.
These parameters are demonstrated in \cref{fig:box_params}.

This parametrization for oriented bounding boxes means that the pixel does not need to be at the center of the box, but does have to be inside the box.
This constraint limits how coarse the bounding box proposals can be to accurately capture entities in the scene.
While having an output resolution of $\sim 1.5 \textrm{m / px}$ is sufficient for vehicles, it would be too coarse to ensure coverage of pedestrians.
We leave a more complex oriented bounding box representation that can accommodate smaller entities without increasing the box resolution  to a future work.

\subsubsection{Training}
Given a set of ground truth boxes $G = \{ g_j \}$ and a set of predicted boxes $B = \{ b_i \}$, we use a matching cost $C(b_i, g_j)$ to define a mapping from each ground truth box $j$ to the best matched predicted box $i(j)$ such that $g_j$ is matched with $b_{i(j)}$.
\cite{sun_what_2021} uses a combination of classification, L1, and IoU costs.
We modify this approach to account for the intractability of IoU for oriented bounding boxes.

The classification cost can be immediately applied to our setting. Our bounding boxes only have one class (vehicles), and so the classification cost is simply the binary cross-entropy between the probability $p(b_i)$ and the probability of $g_j$ (defined to be 1 for all ground truth boxes).

The L1 cost is also easy to apply. For each pair of predicted box and ground truth box we compute the correct box parameters to match the ground truth box from the predicted box's pixel center.
For the log distance terms we clip the distance in cases where the distance is negative (i.e. the pixel is outside the box).
We then apply an L1 loss on the cosine, sine, and four log distance terms of the box representation.

Several works have attempted to approximate IoU for oriented bounding boxes in a differentiable way \cite{kl-rotated-det_2021, gaussian-wasserstein-detection_2021, r3det_2019}. 
Given the simplified detection setting in this work (our input consists of clean, perfect rectangles with unambiguous orientations), we propose a simpler alternative that is tractable for oriented bounding boxes and describes the spatial alignment of the two boxes. We apply a vertex cost term as the average L2 norm between each pair of vertices.
\begin{equation}
C_\textrm{vert} (b_i, g_j) = \frac{1}{4} \sum_{v} || \textrm{vert}_v(b_i) - \textrm{vert}_v(g_j) ||_2
\end{equation}
where $\textrm{vert}_v(b)$ computes the $v^\textrm{th}$ vertex of the box (front left, front right, back left, and back right).
We find that this vertex cost is important in matching, as the L1 loss may prefer a good box for an adjacent vehicle (which gets good L1 loss for all but the left and right distance parameters) over a not-so-good box that aligns spatially with the ground truth box in question.

The final matching cost is a weighted combination of these three costs
\begin{equation}
C = \alpha_\textrm{cls} C_\textrm{cls} + \alpha_\textrm{L1} C_{\textrm{L1}} + \alpha_\textrm{vert} C_\textrm{vert}
\end{equation}

Given these matching costs, we assign one predicted box $b_{i(j)}$ to each ground truth box $g_j$.

A binary cross-entropy loss $\mathcal L_\textrm{cls}$ is used for the predicted box probabilities $p(b_i)$ and an indicator $\mathbbm{1} \left[ \, \exists j : i(j) = i \right]$ for whether $b_i$ was assigned to some ground truth box.
L1 and vertex losses are simply the corresponding matching costs applied between each pair $g_j$ and $b_{i(j)}$. 
Note that the parameters defining box shape are only regressed for predicted boxes that are matched to a ground truth box, while the box logit is trained for all predicted boxes.
The overall detection loss is a weighted combination of the classification, L1, and vertex loss terms. The weighting coefficients do not need to be the same as that used for the matching cost.
\begin{equation}
L_\textrm{rec} = \beta_\textrm{cls} L_\textrm{cls} + \beta_\textrm{L1} L_\textrm{L1} + \beta_\textrm{vert} L_\textrm{vert}
\end{equation}

\section{SCENE DIFFUSION}

After the scene autoencoder is trained, we use the frozen encoder and decoder to train a diffusion model on the latent embeddings as in \cite{latent-diffusion_2022}.
However, for the actual diffusion algorithm we use EDM \cite{karras_elucidating_2022} with the addition of image conditional data.
\cref{fig:diffusion_architecture} shows the architecture for diffusion training and \cref{fig:inference_architecture} shows the architecture for inference.

\begin{figure*}[t]
    \centering
    \includegraphics[width=\linewidth]{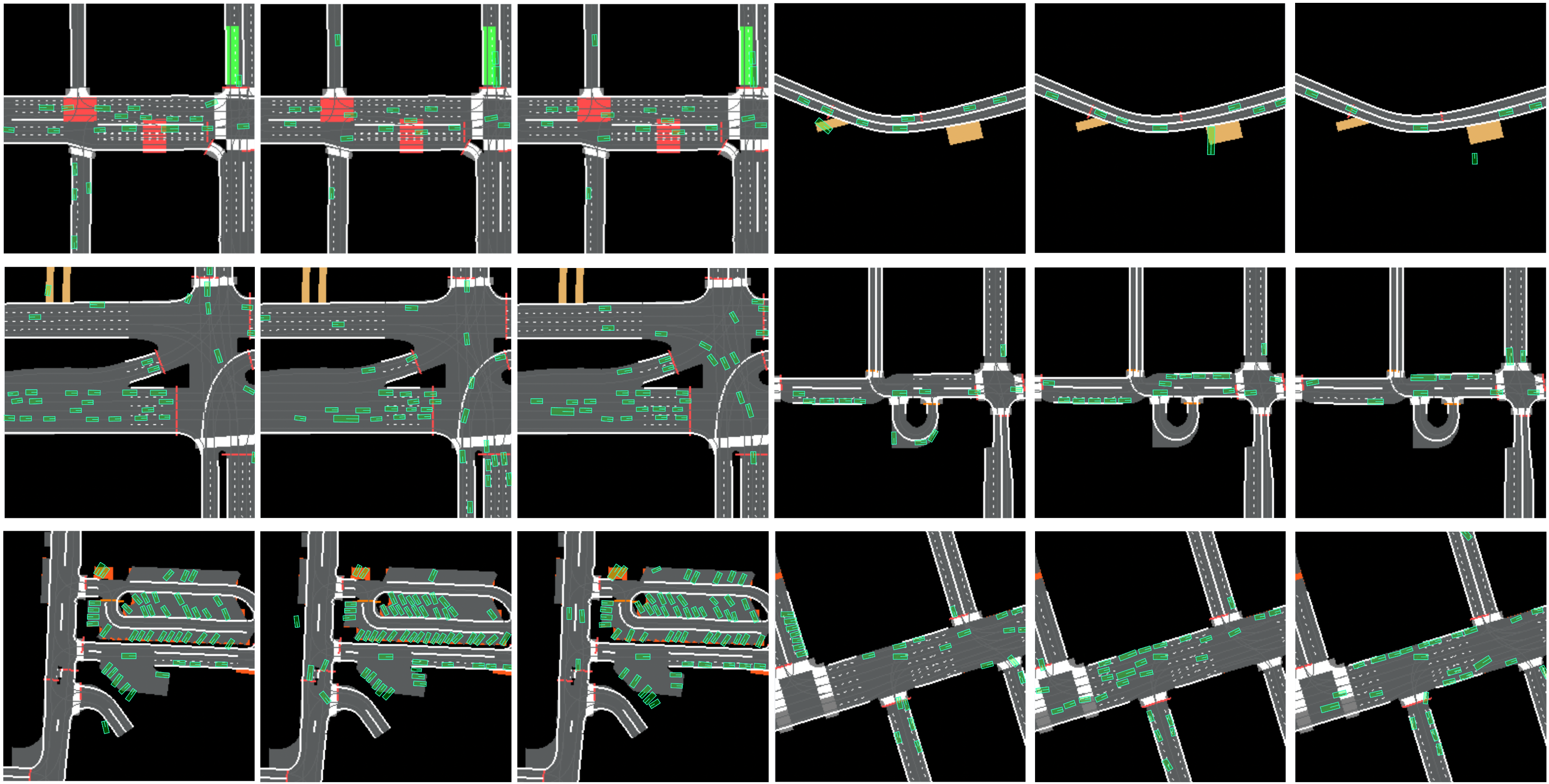}
    \caption{Driving scenes generated from a variety of map locations. A single model trained on the full dataset is able to produce a wide range of scenes. Generating multiple scenes with the same map image and different initial diffusion seeds produces distinct driving situations.}
    \label{fig:gen_scenes}
\end{figure*}

\subsection{Training}

Let $\hat z = \mathcal E (x)$ be a latent embedding obtained from the frozen encoder and $m$ the corresponding map data as described in \cref{sec:problem_setting}.

During training we sample $\sigma$ according to a log normal distribution; i.e. $\log \sigma \sim \mathcal{N} \left( P_\mu, P_\sigma^2 \right)$ with hyperparameters $P_\mu$ and $P_\sigma$.

We create a noisy version of the embeddings $z = \hat z + \sigma \epsilon$ where $\epsilon \in \mathbb{R}^{C' \times H' \times W'}$ is sampled from the standard normal distribution. We then train the denoising model $\mathcal M$ to correct for the noise conditioned on map data $m$ and the noise level $\sigma$ by minimizing the reconstruction loss:
\begin{equation}
\mathcal L_z = \lambda \, c_\textrm{out}^2 \left| \left| \mathcal M \left( c_\textrm{in} \, z; m, \sigma \right) - \frac{\hat z - c_\textrm{skip} \, z}{c_\textrm{out}} \right| \right|_2^2
\end{equation}
where $\lambda$, $c_\textrm{in}$, $c_\textrm{skip}$, and $c_\textrm{out}$ are scalar functions of $\sigma$ as defined in \cite{karras_elucidating_2022}.

Given a noisy sample $z$, we can estimate the denoised version $\hat z$ by 
\begin{equation}
M(z; m, \sigma) = c_\textrm{skip} \, z + c_\textrm{out} \mathcal M \left( c_\textrm{in} \, z; m, \sigma \right)
\end{equation}

The L2 norm on the latent space treats all directions equally, but some will be more or less important for the decoder. To capture this, we apply an additional reconstruction loss
\begin{equation}
\mathcal L_y = \mathcal L_\textrm{rec} \left( \mathcal D \left( M( z; m, \sigma ), m \right) , x\right)
\end{equation}
where $\mathcal D$ is the frozen decoder and $\mathcal L_\textrm{rec}$ is the same reconstruction loss used for the autoencoder in \cref{sec:cvae}.

The final loss for the denoising model is a weighted combination of these two losses:
\begin{equation}
\mathcal L_\textrm{diff} = \mathcal L_z + \beta_y \, \mathcal L_y
\end{equation}

\subsection{Inference}

To generate new samples given a map image $m$, an initial noisy sample $z \sim \mathcal N(0, \sigma_\textrm{max}^2 \mathbf I)$ is drawn for some large noise level $\sigma_\textrm{max}$.
This sample is then iteratively refined according to the reverse process ODE as described in \cite{karras_elucidating_2022}.
Due to the manner in which the denoising model $\mathcal M$ is trained, the gradient of the log probability $\nabla_z \log p(z ; m, \sigma) = \left( M(z; m, \sigma) - z \right) / \sigma^2$. Thus, the ODE simplifies to

\begin{equation} \label{eq:ode}
\frac{\textrm{d} z}{\textrm{d} \sigma} = - \sigma \nabla_{ z} \log p( z ; m, \sigma) = \frac{ z - M( z; m, \sigma)}{\sigma}
\end{equation}

After the sample has been integrated from $\sigma_\text{max}$ to 0, it is passed through the decoder to get the final output $y = \mathcal D(z, m)$.

\section{EXPERIMENTAL RESULTS}
\subsection{Experiment Details}

\subsubsection{Dataset}

\blindSwitch{We use a private dataset containing 6 million driving scenes from San Francisco, Las Vegas, Seattle, and the campus of the Stanford Linear Accelerator Center (SLAC).}{We use a private dataset containing 6 million driving scenes from regions A, B, C, and D.\footnote{The actual names of the city for each data collection region has been anonymized for double-blind review, and will be restored in the accepted paper.} Regions A, B, and C are major metropolitan areas in the US. Region D is a suburban academic campus.}
Scenes are sampled every 0.5 seconds from real-world driving logs.

Each scene contains vehicle tracks as detected by an on-vehicle perception system.
We filter these tracks to only include bounding boxes that overlap with drivable areas in the map (e.g. roads, public parking lots, driveways). 

Scenes are centered on the autonomous vehicle.
The autonomous vehicle is treated as just another vehicle in the scene during training, so the model always sees scenes with a vehicle in the scene center.

The birds' eye view images are $H = W = 256$ pixels in height and width and cover an area 100m x 100m.
The entity image $x$ has $C_x = 3$ channels where the box for each agent is filled with ones, the sine of the agent's heading, and the cosine of the agent's heading, respectively.

\subsubsection{Model Architecture}

In the autoencoder, we use the same encoder architecture as \cite{latent-diffusion_2022} with $f=3$ downsampling layers so that the output latent embedding is $H' = W' = 32$ pixels in height and width.
The encoder and decoder use 32, 64, 128, and 128 channels at the 0x, 1x, 2x, and 3x downsampled feature levels, respectively.
We use a latent channel dimension of $C' = 4$.

For the decoder, we first use the same architecture as the encoder to downsample the map image to the same pixel dimension as the latent embedding.
However, we also save feature maps at each level of downsampling.
We then concatenate the compressed map data with the latent embedding from the encoder and use a modified version of the decoder from \cite{latent-diffusion_2022} where we add skip connections in each upsampling block to the corresponding feature map from the map downsampling network.
The decoder does not upsample all the way to the original encoder input size, and instead outputs bounding box detections at 64 x 64 pixels.
This is a high enough resolution to consistently have a pixel within each vehicle, while low enough to avoid memory issues during training.

In the denoising model we first use the same architecture as the autoencoder's encoder to process the conditional map image down to the same pixel size as the latent embedding.
The processed map data is concatenated with the latent embedding and processed by a time-conditioned Unet as in \cite{ddpm_2020, ddim_2020, latent-diffusion_2022}.
The Unet uses 64, 128, and 256 layers in three levels of features.
At the lowest resolution layer it uses self-attention with 8 heads.
Each resolution layer uses 2 residual blocks.

We threshold the boxes generated by the decoder based on the generated box probabilities.
We use a probability threshold of $90\%$, although we observe that almost all generated boxes have probability either above 97\% or below 40\%.

\subsubsection{Training}
For training the autoencoder, we use the Adam optimizer \cite{adam_2014} with a learning rate of 1e-4 and a weight decay of 1e-5.
For the detection matching we use $\alpha_\textrm{cls} = 4$, $\alpha_\textrm{L1} = 1$, and $\alpha_\textrm{vert} = 1$.
For the detection loss we use $\beta_\textrm{cls} = 20$, $\beta_\textrm{L1} = 1$, and $\beta_\textrm{vert} = 1$.

For training the diffusion model, we use $P_\mu = -0.5$ and $P_\sigma = 1$.
The loss weight $\beta_y = 0.2$.
We use an AdamW optimizer \cite{adamw_2017} with a learning rate of 3e-4 and weight decay of 1e-5.

For inference we use 100 timesteps and select a noise schedule with $\rho = 7$ from \cite{karras_elucidating_2022}.

\begin{table}[t]
    \centering
    \caption{MMD\textsuperscript{2} metric on agent positions evaluated across regions. Models trained on a single region perform the best on scenes from that region. The model trained on all four regions is able to come close to the performance of the region-specific models.}
    \begin{tabular}{c || c | c | c | c}
        \makecell{MMD\textsuperscript{2}\\Position ($\downarrow$)} & \makecell{\LV \\ Scenes} & \makecell{\SEA \\ Scenes} & \makecell{\SF \\ Scenes} & \makecell{\SLAC \\ Scenes} \\ \hline 
        \LV{} Model & \textbf{0.077} & 0.116 & 0.092 & 0.218 \\ 
        \SEA{} Model & 0.112 & \textbf{0.067} & 0.067 & 0.167 \\ 
        \SF{} Model & 0.106 & 0.081 & \textbf{0.042} & 0.158 \\ 
        \SLAC{} Model & 0.151 & 0.129 & 0.113 & \textbf{0.052} \\ \hline 
        Full Model & 0.078 & 0.074 & 0.043 & 0.060 \\
    \end{tabular}
    \label{tab:mmd2_pos}
\end{table}

\begin{table}[t]
    \centering
    \caption{MMD\textsuperscript{2} metric on agent heading evaluated across regions. We observe a similar pattern as the agent position MMD\textsuperscript{2} metric in \cref{tab:mmd2_pos}.}
    \begin{tabular}{c || c | c | c | c}
        \makecell{MMD\textsuperscript{2}\\Heading ($\downarrow$)} & \makecell{\LV \\ Scenes} & \makecell{\SEA \\ Scenes} & \makecell{\SF \\ Scenes} & \makecell{\SLAC \\ Scenes} \\ \hline 
        \LV{} Model & \textbf{0.095} & 0.156 & 0.137 & 0.306 \\
        \SEA{} Model & 0.134 & \textbf{0.064} & 0.086 & 0.246 \\
        \SF{} Model & 0.144 & 0.094 & \textbf{0.037} & 0.245 \\
        \SLAC{} Model & 0.256 & 0.232 & 0.222 & \textbf{0.066} \\ \hline
        Full Model & \textbf{0.095} & 0.076 & 0.040 & 0.078 \\
    \end{tabular}
    \label{tab:mmd2_head}
\end{table}

\subsection{Metrics} \label{sec:metrics}

To determine whether the data produced under our learned models matches the training data used to train the model, we can examine whether the training distribution matches the synthesized data distribution. Since we do not have access to the actual distribution implicit in the diffusion model, we compare the distributions of the samples using the mean maximum discrepancy (MMD) \cite{mmd_2012} under a Gaussian kernel. Given two distributions $p, q$ over $\mathbb{R}^d$,
\begin{equation}
\begin{split}
    \textrm{MMD}^2(p, q) = & \mathbb{E}_{x_1, x_2 \sim p} \left[ k(x_1, x_2) \right] \\
    &+ \mathbb{E}_{y_1, y_2 \sim q} \left[ k(y_1, y_2) \right] \\
    &- 2 \mathbb{E}_{x \sim p, y \sim q} \left[ k(x, y) \right]
\end{split}
\end{equation}
where $k$ is a Gaussian kernel.

We apply this metric to the agent center positions and their heading vectors (i.e. a unit vector in the direction the agent is facing).
We compute this metric between individual pairs of scenes using the same map location and average across the dataset as in \cite{trafficgen_2022}.
This metric quantifies how well the distribution of generated scenes matches the given data sample at the same map location.

\subsection{Generating Driving Scenes}

\cref{fig:gen_scenes} shows a number of generated driving scenes for various map locations.
The model is able to generate diverse driving scenarios for various map locations.
The generated bounding boxes conform to the map and are arranged in a realistic manner.

\subsection{Performance Across Regions}

\begin{figure}[t]
\centering
    \centering
    \includegraphics[width=\linewidth]{\blindSwitch{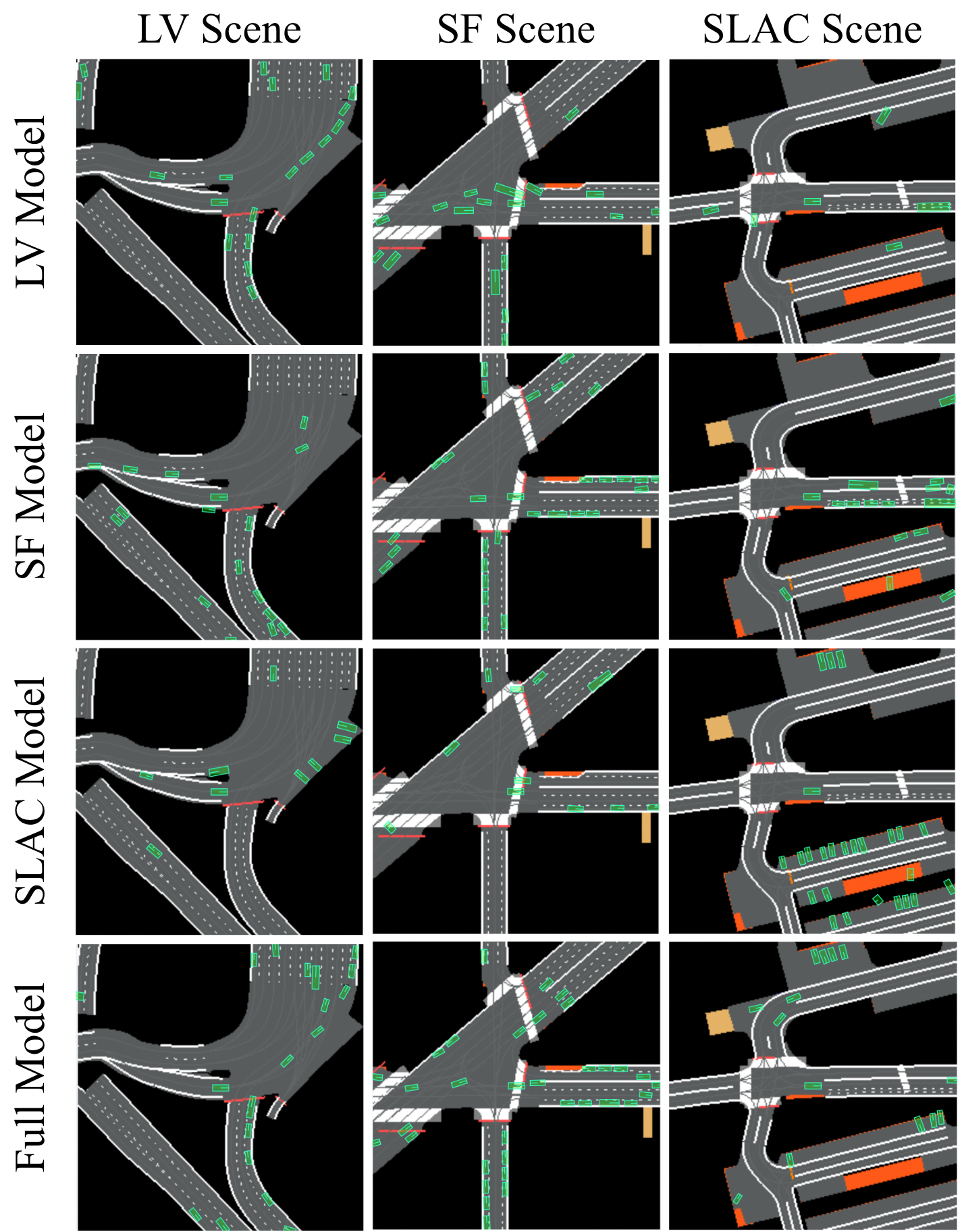}{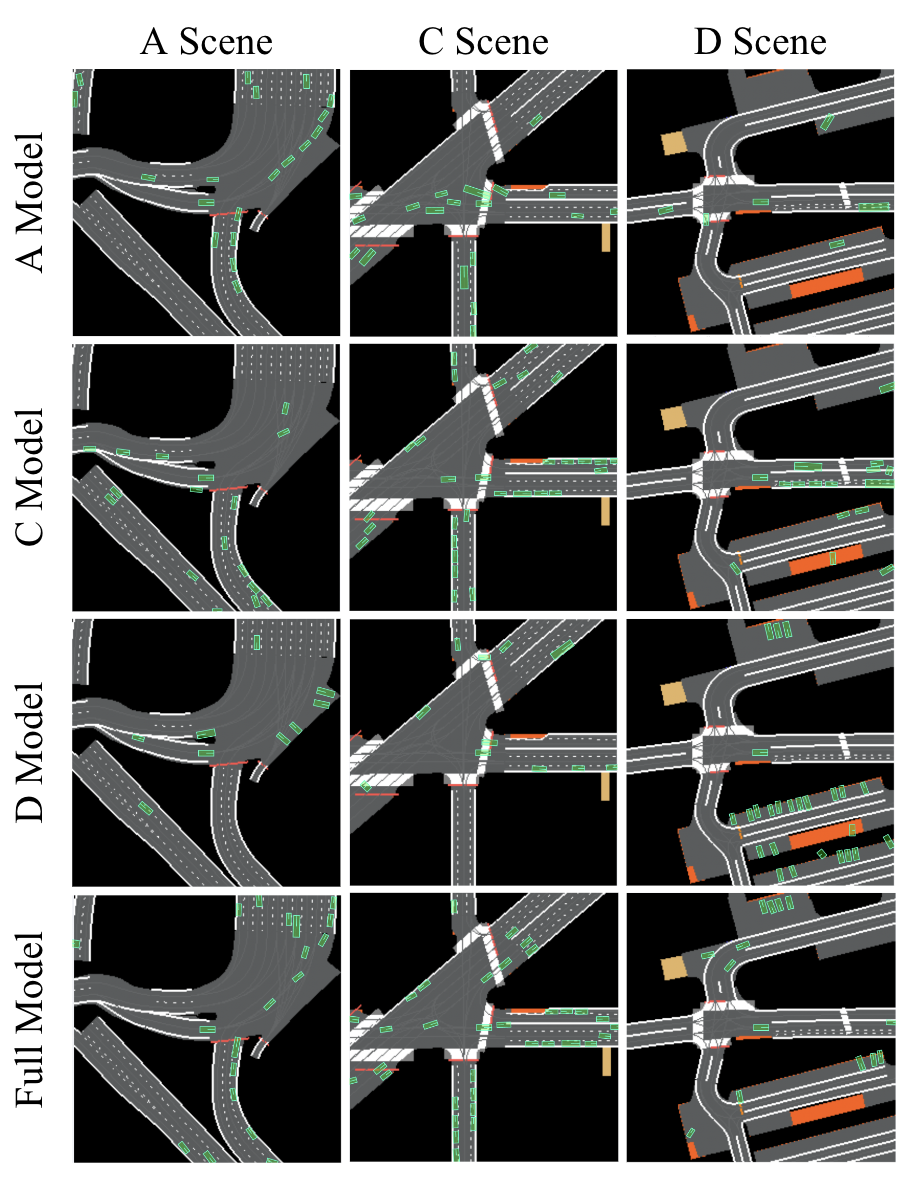}}
    \caption{Driving scenes generated by models trained on data from only one region compared to a model trained on the full dataset. Unique aspects of each region (a line of taxis at a hotel in \LV, parallel parking lanes in \SF, perpendicular parking lots in \SLAC) are only expressed by the models trained on data from that region. However, scenes generated in novel regions are still generally reasonable.}
\label{fig:cross_geofence_samples}
\end{figure}

The four regions included in our dataset each have unique driving features, road geometries, and distributions of vehicles.
To quantify how well our model is able to generalize and learn multiple regions simultaneously, we train the same architecture on data from one region at a time, as well as the full dataset with all four regions.
We then generate scenes using map locations from each region and compute MMD metrics as described in \cref{sec:metrics}.

\cref{tab:mmd2_pos} shows the results for agent positions and \cref{tab:mmd2_head} shows the results for agent headings.
We observe that for both metrics and across all regions the model trained exclusively on that region performs best (i.e. has the lowest MMD).
In all cases the model trained on the full dataset comes close to matching this performance.
This demonstrates that our model has the capacity to incorporate the unique aspects of all four regions in a single model.

The model trained on \blindSwitch{SLAC}{Region D} data consistency performs the worst in the other three regions, and \blindSwitch{SLAC}{Region D} scenes show the largest gap between models trained with and without that data.
This intuitively makes sense, as \blindSwitch{SLAC}{Region D} is an academic campus while the other three regions consist of dense urban scenes.

Qualitatively, we observe that models trained on only one region are unable to capture features of the driving scene unique to another region.
\cref{fig:cross_geofence_samples} compares scenes generated by models trained on different regions.



\section{RELATED WORKS}

Several works have generated agents conditioned on map information using different approaches.
We also provide an overview of some relevant works from diffusion for natural images and oriented object detection.

\subsection{Driving Scene Generation} \label{ssec:generating_scenes}

A number of prior works have used heuristics or fixed grammars to generate driving scenes \cite{izatt2020generative,fisher2012example, carla_2017, metasim_2019}.
Such approaches are limited in their ability to generate large and diverse driving scenes using large-scale datasets.

One learned approach to scene generation is to autoregressively generate one agent at a time.
SceneGen \cite{scenegen_2021} uses a birds' eye view image of the map and a convolutional LSTM to iteratively predict a probability distribution over the next agent, moving from left to right in the scene.
This probability distribution is factorized with chained terms for the object's class, position, heading, extent, and velocity.

TrafficGen \cite{trafficgen_2022} also uses an autoregressive generation process.
They replace the rasterized map image with a vector representation \cite{vectornet_2020}.
They also generate trajectories given the initial agent states with a disjoint second model that adapts \cite{multipath++_2021} to only consider a single timestep of history.

In image synthesis, an alternative approach to autoregressive generation such as \cite{pixelrnn_2016, taming-transformers_2021} is to generate the entire image simultaneously.
SimNet \cite{simnet_2021} uses a conditional GAN \cite{cgan_2016} to generate an occupancy map and apply heuristic post-processing to identify connected components and fit bounding boxes to them.
A second agent-centric model is then used to generate trajectories for these agents over time.

Our work adapts a different technique from image synthesis, latent diffusion \cite{latent-diffusion_2022}, to generate the entire driving scene simultaneously.
In contrast to \cite{simnet_2021}, our model directly outputs oriented bounding boxes for agents and does not require post-processing to fit bounding boxes to occupancy maps. 

\subsection{Diffusion for Image Synthesis} \label{ssec:diffusion_image}

Diffusion models have become quite popular for generating images.
Many works have explored conditioning the diffusion model on other image or text data (e.g. \cite{latent-diffusion_2022}).

\cite{latent-diffusion_2022} uses an autoencoder to compress images to a latent embedding and then applies diffusion to model the data distribution in that latent space.
This decouples the output representation with the diffusion representation, allowing the diffusion model to focus on the semantic structure of the image while the decoder handles perceptual details.

\cite{karras_elucidating_2022} proposes a simplified algorithm for both training and inference using diffusion models. Their approach improves performance in image synthesis and simplifies the hyperparameters required.

\subsection{Object Detection} \label{ssec:obj_detection}

Oriented object detection has been studied in many contexts.
In contrast to other works, our problem setting simplifies the detection task in several ways.
While aerial imagery and text recognition can deal with very large aspect ratios \cite{r3det_2019, kl-rotated-det_2021}, the sizes and aspect ratios of our bounding boxes are more constrained by the dimensions of vehicles.
Furthermore, each vehicle's bounding box has a canonical orientation (i.e. the direction the vehicle is facing), removing the ambiguity over how to define box orientation and the ``angle boundary discontinuity'' present in other works \cite{r3det_2019, kl-rotated-det_2021, pair-midlines_2019, gaussian-wasserstein-detection_2021}.
While 2-stage \cite{rotation-proposals_2017} and non-differentiable \cite{oriented-bbox-rotated_2021} approaches have been explored for other oriented object detection tasks, in this work we desire an end-to-end differentiable approach that does not require extensive post-processing.

\section{CONCLUSIONS}

In this work we presented a novel approach to generating complex driving scenes using diffusion in an end-to-end differentiable architecture that directly generates discrete agents.
We also analyze the generalization capabilities of our model across multiple regions.

Generating diverse, realistic, and complex driving scenarios is a key part of scaling the validation of autonomous driving systems.
We believe this work provides a new approach to address this challenge that is more stable, controllable, and higher quality than prior approaches. In future work, we plan to extend these ideas beyond synthesis of the initial scenario to the time series generation of agents moving in a dynamic scenario. 
We hope this will contribute to the safe deployment of autonomous driving systems in the coming years.


\section*{ACKNOWLEDGMENT}

We thank Gary Linscott, Jake Ware, and Meghana Reddy Ganesina for helpful feedback on the paper; Allan Zelener and Chris Song for discussions on object detection; Peter Schleede and Andres Morales for discussions on diffusion.
We also thank the organizers and reviewers of the ICRA Scalable Autonomous Driving Workshop.


\bibliographystyle{IEEEtran}
\bibliography{IEEEabrv,root}

\begin{thebibliography}{10}
\providecommand{\url}[1]{#1}
\csname url@rmstyle\endcsname
\providecommand{\newblock}{\relax}
\providecommand{\bibinfo}[2]{#2}
\providecommand\BIBentrySTDinterwordspacing{\spaceskip=0pt\relax}
\providecommand\BIBentryALTinterwordstretchfactor{4}
\providecommand\BIBentryALTinterwordspacing{\spaceskip=\fontdimen2\font plus
\BIBentryALTinterwordstretchfactor\fontdimen3\font minus
  \fontdimen4\font\relax}
\providecommand\BIBforeignlanguage[2]{{%
\expandafter\ifx\csname l@#1\endcsname\relax
\typeout{** WARNING: IEEEtran.bst: No hyphenation pattern has been}%
\typeout{** loaded for the language `#1'. Using the pattern for}%
\typeout{** the default language instead.}%
\else
\language=\csname l@#1\endcsname
\fi
#2}}

\bibitem{Brooks1993}
R.~A. Brooks and M.~J. Mataric, ``Real robots, real learning problems,'' in
  \emph{Robot Learning}, J.~H. Connell and S.~Mahadevan, Eds.\hskip 1em plus
  0.5em minus 0.4em\relax Boston, MA: Springer US, 1993, pp. 193--213.

\bibitem{izatt2020generative}
G.~Izatt and R.~Tedrake, ``Generative modeling of environments with scene
  grammars and variational inference,'' in \emph{2020 IEEE International
  Conference on Robotics and Automation (ICRA)}.\hskip 1em plus 0.5em minus
  0.4em\relax IEEE, 2020, pp. 6891--6897.

\bibitem{fisher2012example}
M.~Fisher, D.~Ritchie, M.~Savva, T.~Funkhouser, and P.~Hanrahan,
  ``Example-based synthesis of 3d object arrangements,'' \emph{ACM Transactions
  on Graphics (TOG)}, vol.~31, no.~6, pp. 1--11, 2012.

\bibitem{scenegen_2021}
\BIBentryALTinterwordspacing
S.~Tan, K.~Wong, S.~Wang, S.~Manivasagam, M.~Ren, and R.~Urtasun, ``Scenegen:
  Learning to generate realistic traffic scenes,'' \emph{CVPR}, 2021. [Online].
  Available: \url{https://arxiv.org/abs/2101.06541}
\BIBentrySTDinterwordspacing

\bibitem{trafficgen_2022}
\BIBentryALTinterwordspacing
L.~Feng, Q.~Li, Z.~Peng, S.~Tan, and B.~Zhou, ``{TrafficGen}: {Learning} to
  {Generate} {Diverse} and {Realistic} {Traffic} {Scenarios},'' Oct. 2022.
  [Online]. Available: \url{http://arxiv.org/abs/2210.06609}
\BIBentrySTDinterwordspacing

\bibitem{simnet_2021}
\BIBentryALTinterwordspacing
L.~Bergamini, Y.~Ye, O.~Scheel, L.~Chen, C.~Hu, L.~Del~Pero, B.~Osinski,
  H.~Grimmett, and P.~Ondruska, ``{SimNet}: {Learning} {Reactive}
  {Self}-driving {Simulations} from {Real}-world {Observations},'' May 2021.
  [Online]. Available: \url{http://arxiv.org/abs/2105.12332}
\BIBentrySTDinterwordspacing

\bibitem{pixelrnn_2016}
\BIBentryALTinterwordspacing
A.~v.~d. Oord, N.~Kalchbrenner, and K.~Kavukcuoglu, ``Pixel recurrent neural
  networks,'' 2016. [Online]. Available: \url{https://arxiv.org/abs/1601.06759}
\BIBentrySTDinterwordspacing

\bibitem{cgan_2016}
\BIBentryALTinterwordspacing
P.~Isola, J.-Y. Zhu, T.~Zhou, and A.~A. Efros, ``Image-to-image translation
  with conditional adversarial networks,'' 2016. [Online]. Available:
  \url{https://arxiv.org/abs/1611.07004}
\BIBentrySTDinterwordspacing

\bibitem{ddpm_2020}
\BIBentryALTinterwordspacing
J.~Ho, A.~Jain, and P.~Abbeel, ``Denoising diffusion probabilistic models,''
  2020. [Online]. Available: \url{https://arxiv.org/abs/2006.11239}
\BIBentrySTDinterwordspacing

\bibitem{diffusion-beats-gans_2021}
\BIBentryALTinterwordspacing
P.~Dhariwal and A.~Nichol, ``Diffusion {Models} {Beat} {GANs} on {Image}
  {Synthesis},'' June 2021. [Online]. Available:
  \url{http://arxiv.org/abs/2105.05233}
\BIBentrySTDinterwordspacing

\bibitem{latent-diffusion_2022}
\BIBentryALTinterwordspacing
R.~Rombach, A.~Blattmann, D.~Lorenz, P.~Esser, and B.~Ommer, ``High-resolution
  image synthesis with latent diffusion models,'' \emph{CVPR}, 2021. [Online].
  Available: \url{https://arxiv.org/abs/2112.10752}
\BIBentrySTDinterwordspacing

\bibitem{rules-of-road_2019}
\BIBentryALTinterwordspacing
J.~Hong, B.~Sapp, and J.~Philbin, ``Rules of the road: Predicting driving
  behavior with a convolutional model of semantic interactions,'' \emph{CVPR},
  2019. [Online]. Available: \url{http://arxiv.org/abs/1906.08945}
\BIBentrySTDinterwordspacing

\bibitem{kingma_variational-bayes_2013}
\BIBentryALTinterwordspacing
D.~P. Kingma and M.~Welling, ``Auto-encoding variational bayes,'' 2013.
  [Online]. Available: \url{https://arxiv.org/abs/1312.6114}
\BIBentrySTDinterwordspacing

\bibitem{higgins_beta-vae_2022}
\BIBentryALTinterwordspacing
I.~Higgins, L.~Matthey, A.~Pal, C.~Burgess, X.~Glorot, M.~Botvinick,
  S.~Mohamed, and A.~Lerchner, ``\BIBforeignlanguage{en}{beta-{VAE}: {Learning}
  {Basic} {Visual} {Concepts} with a {Constrained} {Variational}
  {Framework}},'' Nov. 2016. [Online]. Available:
  \url{https://openreview.net/forum?id=Sy2fzU9gl}
\BIBentrySTDinterwordspacing

\bibitem{sun_what_2021}
\BIBentryALTinterwordspacing
P.~Sun, Y.~Jiang, E.~Xie, Z.~Yuan, C.~Wang, and P.~Luo, ``What makes for
  end-to-end object detection?'' \emph{ICML}, 2020. [Online]. Available:
  \url{https://arxiv.org/abs/2012.05780}
\BIBentrySTDinterwordspacing

\bibitem{oriented-bbox-rotated_2021}
\BIBentryALTinterwordspacing
M.~Zand, A.~Etemad, and M.~A. Greenspan, ``Oriented bounding boxes for small
  and freely rotated objects,'' \emph{IEEE Transactions on Geoscience and
  Remote Sensing}, 2021. [Online]. Available:
  \url{https://arxiv.org/abs/2104.11854}
\BIBentrySTDinterwordspacing

\bibitem{r3det_2019}
\BIBentryALTinterwordspacing
X.~Yang, Q.~Liu, J.~Yan, and A.~Li, ``R3det: Refined single-stage detector with
  feature refinement for rotating object,'' 2019. [Online]. Available:
  \url{http://arxiv.org/abs/1908.05612}
\BIBentrySTDinterwordspacing

\bibitem{gaussian-wasserstein-detection_2021}
\BIBentryALTinterwordspacing
X.~Yang, J.~Yan, Q.~Ming, W.~Wang, X.~Zhang, and Q.~Tian, ``Rethinking rotated
  object detection with gaussian wasserstein distance loss,'' 2021. [Online].
  Available: \url{https://arxiv.org/abs/2101.11952}
\BIBentrySTDinterwordspacing

\bibitem{kl-rotated-det_2021}
\BIBentryALTinterwordspacing
X.~Yang, X.~Yang, J.~Yang, Q.~Ming, W.~Wang, Q.~Tian, and J.~Yan, ``Learning
  high-precision bounding box for rotated object detection via kullback-leibler
  divergence,'' 2021. [Online]. Available:
  \url{https://arxiv.org/abs/2106.01883}
\BIBentrySTDinterwordspacing

\bibitem{karras_elucidating_2022}
\BIBentryALTinterwordspacing
T.~Karras, M.~Aittala, T.~Aila, and S.~Laine, ``Elucidating the {Design}
  {Space} of {Diffusion}-{Based} {Generative} {Models},'' 2022. [Online].
  Available: \url{http://arxiv.org/abs/2206.00364}
\BIBentrySTDinterwordspacing

\bibitem{ddim_2020}
\BIBentryALTinterwordspacing
J.~Song, C.~Meng, and S.~Ermon, ``Denoising diffusion implicit models,'' 2020.
  [Online]. Available: \url{https://arxiv.org/abs/2010.02502}
\BIBentrySTDinterwordspacing

\bibitem{adam_2014}
\BIBentryALTinterwordspacing
D.~P. Kingma and J.~Ba, ``Adam: A method for stochastic optimization,'' 2014.
  [Online]. Available: \url{https://arxiv.org/abs/1412.6980}
\BIBentrySTDinterwordspacing

\bibitem{adamw_2017}
\BIBentryALTinterwordspacing
I.~Loshchilov and F.~Hutter, ``Fixing weight decay regularization in adam,''
  \emph{ICLR}, 2017. [Online]. Available: \url{http://arxiv.org/abs/1711.05101}
\BIBentrySTDinterwordspacing

\bibitem{mmd_2012}
\BIBentryALTinterwordspacing
A.~Gretton, K.~M. Borgwardt, M.~J. Rasch, B.~Sch{{\"o}}lkopf, and A.~Smola, ``A
  kernel two-sample test,'' \emph{Journal of Machine Learning Research},
  vol.~13, no.~25, pp. 723--773, 2012. [Online]. Available:
  \url{http://jmlr.org/papers/v13/gretton12a.html}
\BIBentrySTDinterwordspacing

\bibitem{carla_2017}
\BIBentryALTinterwordspacing
A.~Dosovitskiy, G.~Ros, F.~Codevilla, A.~M. L{\'{o}}pez, and V.~Koltun,
  ``{CARLA:} an open urban driving simulator,'' 2017. [Online]. Available:
  \url{http://arxiv.org/abs/1711.03938}
\BIBentrySTDinterwordspacing

\bibitem{metasim_2019}
\BIBentryALTinterwordspacing
A.~Kar, A.~Prakash, M.~Liu, E.~Cameracci, J.~Yuan, M.~Rusiniak, D.~Acuna,
  A.~Torralba, and S.~Fidler, ``Meta-sim: Learning to generate synthetic
  datasets,'' 2019. [Online]. Available: \url{http://arxiv.org/abs/1904.11621}
\BIBentrySTDinterwordspacing

\bibitem{vectornet_2020}
\BIBentryALTinterwordspacing
J.~Gao, C.~Sun, H.~Zhao, Y.~Shen, D.~Anguelov, C.~Li, and C.~Schmid,
  ``Vectornet: Encoding {HD} maps and agent dynamics from vectorized
  representation,'' 2020. [Online]. Available:
  \url{https://arxiv.org/abs/2005.04259}
\BIBentrySTDinterwordspacing

\bibitem{multipath++_2021}
\BIBentryALTinterwordspacing
B.~Varadarajan, A.~Hefny, A.~Srivastava, K.~S. Refaat, N.~Nayakanti,
  A.~Cornman, K.~Chen, B.~Douillard, C.~Lam, D.~Anguelov, and B.~Sapp,
  ``Multipath++: Efficient information fusion and trajectory aggregation for
  behavior prediction,'' \emph{ICRA}, 2021. [Online]. Available:
  \url{https://arxiv.org/abs/2111.14973}
\BIBentrySTDinterwordspacing

\bibitem{taming-transformers_2021}
\BIBentryALTinterwordspacing
P.~Esser, R.~Rombach, and B.~Ommer, ``Taming {Transformers} for
  {High}-{Resolution} {Image} {Synthesis},'' 2020. [Online]. Available:
  \url{http://arxiv.org/abs/2012.09841}
\BIBentrySTDinterwordspacing

\bibitem{pair-midlines_2019}
\BIBentryALTinterwordspacing
H.~Wei, L.~Zhou, Y.~Zhang, H.~Li, R.~Guo, and H.~Wang, ``Oriented objects as
  pairs of middle lines,'' \emph{ISPRS Journal of Photogrammetry and Remote
  Sensing}, 2019. [Online]. Available: \url{http://arxiv.org/abs/1912.10694}
\BIBentrySTDinterwordspacing

\bibitem{rotation-proposals_2017}
\BIBentryALTinterwordspacing
J.~Ma, W.~Shao, H.~Ye, L.~Wang, H.~Wang, Y.~Zheng, and X.~Xue,
  ``Arbitrary-oriented scene text detection via rotation proposals,'' 2017.
  [Online]. Available: \url{http://arxiv.org/abs/1703.01086}
\BIBentrySTDinterwordspacing

\end{thebibliography}

\end{document}